\documentclass[letterpaper, 10 pt, conference]{ieeeconf}  

\IEEEoverridecommandlockouts                              

\usepackage{graphics} 
\usepackage{epsfig} 
\usepackage{mathptmx} 
\usepackage{times} 
\usepackage{amsmath} 
\usepackage{amssymb}  
\usepackage{xcolor}
\usepackage{graphicx}
\usepackage{url}
\usepackage[hidelinks]{hyperref}

\usepackage{subcaption}

\usepackage{tikz}
\usepackage{tikz}
\usetikzlibrary{overlay-beamer-styles}
\usetikzlibrary{backgrounds}
\usetikzlibrary{positioning}
\usetikzlibrary{arrows.meta}
\usetikzlibrary{tikzmark}
\usetikzlibrary{shapes}
\usetikzlibrary{calc}
\usepackage{pgfplots}
\pgfplotsset{compat=1.16} 

\definecolor{bluecarlo}{rgb}{.35, 0.51, 0.69}
\definecolor{redcarlo}{rgb}{1., 0.43, 0.43}
\definecolor{yellowcarlo}{rgb}{1., 0.83, 0.15}

 \tikzset{
    highlightred on/.style={alt={#1{color=red}{}}},
    highlightorange on/.style={alt={#1{color=orange}{}}},
    highlightgreen on/.style={alt={#1{color=green}{}}},
    highlightbold on/.style={alt={#1{font=\bfseries}{}}}
}

\newcommand{\rebuttalOne}[1]{\textcolor{black}{#1}}
\newcommand{\rebuttalThree}[1]{\textcolor{black}{#1}}
\newcommand{\rebuttalSix}[1]{\textcolor{black}{#1}}
\newcommand{\finalSentence}[1]{\textcolor{black}{#1}}

\title{\LARGE \bf
Adaptive Control based Friction Estimation for \\ Tracking Control of Robot Manipulators
}

\author{Junning Huang$^{1}$, Davide Tateo$^{1}$, Puze Liu$^{1, 2}$, Jan Peters$^{1, 2, 3}$
\thanks{$^{1}$Department of Computer Science, Technische Universität Darmstadt
        {\tt\small \{junning, davide, puze\}@robot-learning.de}, {\tt\small jan.peters@tu-darmstadt.de}}%
\thanks{$^{2}$Department of Systems AI for Robot
Learning, German Research Center for AI (DFKI)}%
\thanks{$^{3}$Hessian Centre for Artificial Intelligence}
}

\begin{document}

\let\oldtwocolumn\twocolumn
\renewcommand\twocolumn[1][]{%
    \oldtwocolumn[{#1}{
    \begin{center}
    \vspace{-1.7em}
    \captionsetup{type=figure}
    \includegraphics[width=\textwidth, height=0.3\textwidth]{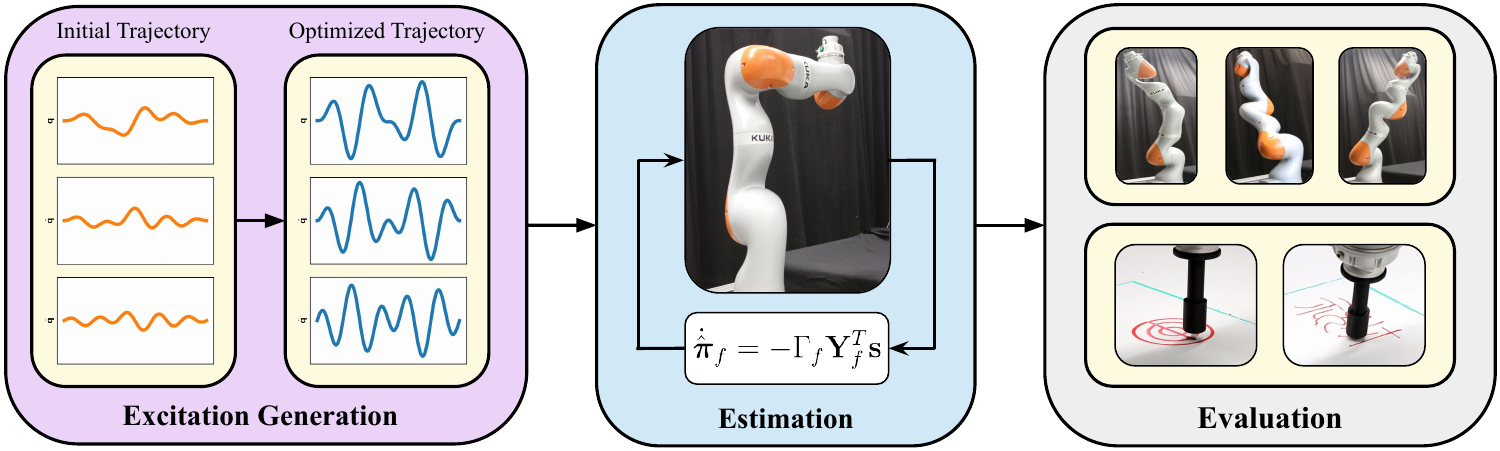}
    \captionof{figure}{Proposed pipeline for friction estimation and evaluation: (\textbf{Left}) we start by generating excitation by solving an optimization problem with random initial trajectories; (\textbf{Center}) an adaptive controller is then proposed to estimate the friction parameters with the excitation we generated from the previous step; (\textbf{Right}) the estimated parameters are evaluated on two trajectory tracking tasks: joint space random Fourier trajectories and cartesian space drawing tasks.} 
    \vspace{0.1em}
    \label{fig:main_figure}
        \end{center}
    }]
}

\maketitle
\thispagestyle{empty}
\pagestyle{empty}

\begin{abstract}
Adaptive control is often used for friction compensation in trajectory tracking tasks because it does not require torque sensors. However, it has some drawbacks: first, the most common certainty-equivalence adaptive control design is based on linearized parameterization of the friction model, therefore nonlinear effects, including the stiction and Stribeck effect, are usually omitted. Second, the adaptive control-based estimation can be biased due to non-zero steady-state error. Third, neglecting unknown model mismatch could result in non-robust estimation.
This paper proposes a novel linear parameterized friction model capturing the nonlinear static friction phenomenon. Subsequently, an adaptive control-based friction estimator is proposed to reduce the bias during estimation based on backstepping. Finally, we propose an algorithm to generate excitation for robust estimation.
Using a KUKA iiwa 14, we conducted trajectory tracking experiments to evaluate the estimated friction model, including random Fourier and drawing trajectories, showing the effectiveness of our methodology in different control schemes. Additional experiments can be found in: \url{https://sites.google.com/view/ral-friction-estimation/home}.
\end{abstract}

\section{INTRODUCTION}
Friction compensation is inherently challenging for two reasons: First, friction models can not be derived from first-principles physics laws due to the highly nonlinear and dynamic micro and macro friction phenomenon. Second, friction is difficult to measure, and most robot manipulators do not have expensive torque sensors. Third, friction estimation is affected by the precision of other dynamic parameters of the robot manipulator, e.g., gravity and inertia.

Empirical models are essential for addressing friction modeling challenges and are categorized into static and dynamic types \cite{olsson1998friction}. Static models, like the nonlinear Stribeck model, efficiently capture nonlinear effects such as the Stribeck effect and static friction, making them suitable for real-time applications. However, they fail to represent dynamic behaviors like hysteresis and stick-slip motion. In practice, to construct a simple model for better parameter estimation, some nonlinear effects of friction are neglected, and only Coulomb and viscous friction are included \cite{armstrong1991control}.

While dynamic models can capture time-dependent behaviors such as hysteresis, they are computationally intensive and often involve complex parameter estimation. They can also be numerically unstable due to stiff ordinary differential equations. These issues make dynamic models less practical, leading to the more frequent use of static models where simplicity and computational efficiency are prioritized.

Adaptive control-based friction estimation or friction observers can estimate friction without torque sensors. The most popular design scheme for adaptive control is the certainty equivalence principle (CE). CE separates the design of control and estimation laws and can guarantee stability with a weak Lyapunov function. However, the CE-based design usually relies on a linear parameterization of the friction model and cannot guarantee parameter convergence. Hence, simplified linear models that omit nonlinear effects from static friction or stribeck effect and relay-based design are usually used for model representation. 

The third challenge, often overlooked in friction estimation, is the unknown model mismatch, which arise from the under modeling of friction or inaccuracies in rigid body dynamics parameters, can severely damage the controller's performance. There are usually two ways to reduce the model mismatch: either learn a residual model as hindsight correction or design optimal excitations for robust parameter estimation. Learning-based methods \cite{reuss2022end, gruenstein2021residual} can learn residual models to compensate for model mismatch, but they only offer hindsight correction, meaning the mismatch still affects the estimation process. Optimal excitation generation can generate informative trajectories to reduce the model uncertainty, but the connection between model mismatch and trajectories in estimation is unclear.

The contributions of this paper are three-fold. We propose (i). A linear parameterized friction model considering static friction and the Stribeck effect allows a simple adaptive control design to estimate friction; (ii). A robust adaptive control-based friction estimator based on input-to-state stability;  (iii). An optimization-based approach for excitation generation that reduces the disturbance during estimation, increasing the estimation process's robustness.

\subsection*{Related Work}
Empirical friction models, including static and dynamic friction models, are proposed to capture the static and dynamic characteristics of friction. The most popular static friction model is the Stribeck model \cite{olsson1998friction}, which includes the phenomenon of Coulomb, Stribeck, and viscous friction. In the early 1990s, dynamic friction models, including the LuGre model \cite{de1995new} and the Maxwell-slip model \cite{al2005generalized}, were proposed to capture the dynamic characteristics of friction, such as the hysteresis phenomenon of friction.
Recently, precise modelings of friction are proposed considering the effect elasticity of joints \cite{wolf2018extending}, or the driving of the robot \cite{lange2021friction}, load and temperature dependence \cite{iskandar2019dynamic}.

Adaptively estimating robot dynamic parameters dates back to \cite{slotine1987adaptive}, where the authors propose a CE-based adaptive control design to estimate the dynamic parameters of a robot manipulator. Later on, the adaptive control design was extended to estimate the friction parameters of the robot, such as \cite{canudas1996adaptive}, \cite{verbert2015adaptive}, and \cite{le2012adaptive}, the authors proposed adaptive controllers to estimate the parameters of a dynamic and a static friction model, respectively. Similar techniques are also used in \cite{panteley1998adaptive} where friction is considered as disturbance and a relay-based adaptive controller is proposed to reject the disturbance.
Friction observer-based compensation is also proposed to estimate the friction parameters \cite{henrichfreise1997observer, le2008friction, kim2019model, putra2004observer, olsson1996observer}, the authors use static or dynamic models as nominal models and design linear Luenberger observer to dynamically estimate the parameters.

Learning residual models \cite{reuss2022end, gruenstein2021residual} to account for a model mismatch to improve the hindsight performance of the estimated model has become popular recently. Different criteria for optimal excitation generation are proposed \cite{swevers1997optimal, lee2021optimal, sturz2017parameter} to retrieve informative trajectories for increasing the robustness of parameter estimation.

\section{PRELIMINARIES}
In this section, we introduce the necessary background for the paper and briefly illustrate the equations of motion for robot manipulators, the empirical friction models, and adaptive control for robot manipulators.

\subsection{Equations of Motion for Robot Manipulators}
The inverse dynamics of a fully-actuated robot manipulators, including friction modeling is given by
\begin{align*}
    \boldsymbol{\tau}   & = \mathbf{M}(\mathbf{q})\ddot{\mathbf{q}}+\mathbf{C}(\mathbf{q}, \dot{\mathbf{q}})\dot{\mathbf{q}}
    + \mathbf{g}(\mathbf{q})+\boldsymbol{\tau}_{f},
\end{align*}
with  $\mathbf{q}\in\mathbb{R}^{n_j}$ is the position in joint space, $\mathbf{M}\in\mathbb{R}^{n_j\times n_j}$ is the joint space inertia matrix,
$\mathbf{C}\in\mathbb{R}^{n_j\times n_j}$ represents the centrifugal and Coriolis matrix, $\mathbf{g}\in\mathbb{R}^{n_j}$ represents gravity, $\boldsymbol{\tau}_{f}$ denotes the torques produced by friction. Here we denote the the number of actuated joints as $n_j$.

It is commonly known that the inverse dynamics of the robot can be represented linearly as the system \cite{wensing2017linear, atkeson1986estimation}
\begin{equation*}
    \mathbf{M}(\mathbf{q})\ddot{\mathbf{q}}+\mathbf{C}(\mathbf{q}, \dot{\mathbf{q}})\dot{\mathbf{q}}
    + \mathbf{g}(\mathbf{q})=\mathbf{Y}(\mathbf{q}, \dot{\mathbf{q}},
    \ddot{\mathbf{q}}) \boldsymbol{\pi},
\end{equation*}
with the regressor matrix $\mathbf{Y}$ and inertial parameters $\pi \in \mathbb{R}^{10n_j}$.


\subsection{Structural Properties of Robot Dynamics}\label{sec:structural_properties}
Robot manipulators belong to the Euler-Lagrange (EL) family, which possess structural properties as follows \cite{munoz2022high}.

\textit{Property 1:} The joint space inertia matrix $\mathbf{M}(\mathbf{q})$ is positive definite and bounded from above and below \cite{munoz2022high}, namely, the eigen value of the joint space inertia matrix satisfies
\mbox{$\sigma_{\min}<\sigma(\mathbf{M}(\mathbf{q}))<\sigma_{\max},$} where $\sigma_{\min}$ and $\sigma_{\max}$ are the minimum and maximum eigenvalues of the joint space inertia matrix, respectively.

\textit{Property 2:} The centrifugal and Coriolis matrix $\mathbf{C}(\mathbf{q}, \dot{\mathbf{q}})$ satisfies
\mbox{$|\mathbf{C}(\mathbf{q}, \dot{\mathbf{q}}_1)\dot{\mathbf{q}}_2|\leq c_0 |\dot{\mathbf{q}}_1| |\dot{\mathbf{q}}_2|,$} where $c_0$ is a positive constant.

\textit{Property 3:} The gravity term $\mathbf{g}(\mathbf{q})$ can be bounded by a positive constant $c_1$, namely, $||\mathbf{g}(\mathbf{q})||\leq c_1$.
These properties can subsequently be used to derive the following proposition.

\textit{Proposition 1:} Due to the kinematic limit of the robot, which results in boundedness of joint position, velocities, and accelerations, the linear regressor is also bounded $||\mathbf{Y}||\leq c_2$, where $c_2$ is a positive constant.

\subsection{Adaptive Control of Robot Manipulators}
Most of the adaptive control design for trajectory tracking tasks of robot manipulators is based on the certainty equivalence principle (CE) adaptive updating law proposed in \cite{slotine1987adaptive}. The CE based control and parameter update laws are 
\begin{align*}
    \tau  =\mathbf{Y}\hat{\boldsymbol{\pi}}-\mathbf{K}_D\mathbf{s}, & \quad &
    \dot{\hat{{\boldsymbol{\pi}}}} =-\Gamma\mathbf{Y}^T\mathbf{s},
\end{align*}
with the tracking error defined on the sliding surface $\mathbf{s}=\dot{\tilde{\mathbf{q}}}+\Sigma\tilde{\mathbf{q}}$, the position $\tilde{\mathbf{q}}=\mathbf{q}-\mathbf{q}_d$, and velocity tracking error $\dot{\tilde{\mathbf{q}}}=\dot{\mathbf{q}}-\dot{\mathbf{q}}_d$, the desired position $\mathbf{q}_d(t)$, a gain matrix $\Sigma$ to determine the sliding surface, a gain matrix $\Gamma$ for the parameter updating law and a gain matrix $\mathbf{K}_D$ for the sliding mode tracking error. In \cite{slotine1987adaptive}, the authors also discussed the case only updating part of the parameters, showing that an additional sliding mode controller $\tau=\mathbf{Y}\hat{\boldsymbol{\pi}}-\mathbf{K}_D\mathbf{s}-\mathbf{k}\text{sgn}(\mathbf{s})$ can guarantee the asymptotic tracking performance.

\section{ADAPTIVE FRICTION ESTIMATION}
In this section, we first define the adaptive estimation problem for friction models.
Subsequently, a friction model including the nonlinear effect and an adaptive friction estimator that estimates the parameters of the proposed friction model is proposed.
The proposed estimator is designed based on the CE principle, and we utilize a simple backstepping design to reduce the bias due to non-zero steady-state error during estimation.
Finally, we discuss the input-to-state stability nature of the proposed estimator and introduce an algorithm for generating excitations, which is crucial for robust estimation.

\subsection{Problem Statement}\label{sec:ps}
In this paper we estimate only the friction parameters $\boldsymbol{\pi}_f$ assuming the estimated rigid body dynamic parameters $\hat{\boldsymbol{\pi}}$ are known, the target is to design an adaptive output feedback tracking controller that ensures 
   $\lim_{t\to\infty}\tilde{\mathbf{x}}=0$,
where
$\tilde{\mathbf{x}}=\begin{bmatrix}
        \tilde{\mathbf{q}} & \dot{\tilde{\mathbf{q}}}
    \end{bmatrix}$, $\tilde{\mathbf{q}}=\mathbf{q}-\mathbf{q}_d$ and $\dot{\tilde{\mathbf{q}}}=\dot{\mathbf{q}}-\dot{\mathbf{q}}_d$. $\mathbf{q}_d(t)$ represent the tracking error of the trajectory, with $\mathbf{q}, \dot{\mathbf{q}}$ the current joint positions and velocities and $\mathbf{q}_d, \dot{\mathbf{q}}_d$ the desired joint positions and velocities.
The proposed controller supposed to be robust to the model mismatch of the rigid body dynamics $\tilde{\boldsymbol{\pi}}$ and friction model $\tilde{\boldsymbol{\pi}}_f$, where $\tilde{\boldsymbol{\pi}}=\hat{\boldsymbol{\pi}}-\boldsymbol{\pi}$ and $\tilde{\boldsymbol{\pi}}_f=\hat{\boldsymbol{\pi}}_f-\boldsymbol{\pi}_f$.

\subsection{Linear Parameterized Friction Model}
Inspired by the friction model in \cite{olsson1998friction}, we propose a linear parameterized friction model that captures the nonlinear effect including stiction and Stribeck effect. We first recall the classic Stribeck friction model \cite{olsson1998friction} is the following
\begin{equation*}
    \tau_f =\sqrt{2e}\left(f_{\text{brk}}-f_{\text{c}}\right)\exp\left(-\frac{v}{v_{\text{st}}}\right)\frac{v}{v_{\text{st}}}+f_{\text{c}} \tanh\left(\frac{v}{v_{\text{coul}}}\right)+f_{\text{vis}}v
\end{equation*}
with $f_{\text{brk}}$ the break-away, $f_{\text{c}}$ the coulomb, and $f_{\text{vis}}$, the viscous friction. $v_{\text{st}}$ is the Stribeck velocity, it determines the timing for compensating the stribeck effect and stiction while $v_{\text{coul}}$ is the velocity determines the start of compensation for coulomb, and $v$ is the relative velocity between the two links, namely the joint velocity. The Stribeck friction model can capture the static characteristics of friction, but it is a nonlinear model and imcompatible with the CE based adaptive control design.

By regarding the coulomb $v_{\text{coul}}$ and Stribeck $v_{\text{st}}$ velocities as tunable parameters, we proposed the following linear regressor
$$\mathbf{Y}_f(v) = \begin{bmatrix}
        \exp(-\frac{v}{v_{\text{st}}})\frac{v}{v_{\text{st}}} & \tanh(\frac{v}{v_{\text{coul}}}) & v \\
    \end{bmatrix} .$$
The above friction model can then be rewritten as a linear model w.r.t. the regressor as
$\tau_f = \mathbf{Y}_f(v)\boldsymbol{\pi}_f$,
where $\boldsymbol{\pi}_f=\begin{bmatrix}f_{\text{brk}}-f_{\text{c}} & f_{\text{c}} & f_{\text{vis}}\end{bmatrix}^T$.
The linearity of the model allows the design of adaptive controllers for parameter estimation. Also, by setting $v_{\text{st}}$ and $v_{\text{coul}}$, we can control the timing of compensation of the static and Coulomb friction, which is a nice feature shared with the classic Stribeck friction model. To obtain reasonable values of $v_{\text{st}}$ and $v_{\text{coul}}$, one can parameterize them as $v_{\text{st}}=v_{\text{brk}}\sqrt{2}$ and $v_{\text{coul}}=v_{\text{brk}}/10$ \cite{armstrong1991control}. Here $v_{\text{brk}}$ is the breakaway velocity, determine the velocity threshold for the robot to move. By analyzing a tracking trajectory's joint position, one can estimate $v_{\text{brk}}$.

\begin{figure}[t]
    \centering
    \includegraphics[width=0.8\columnwidth]{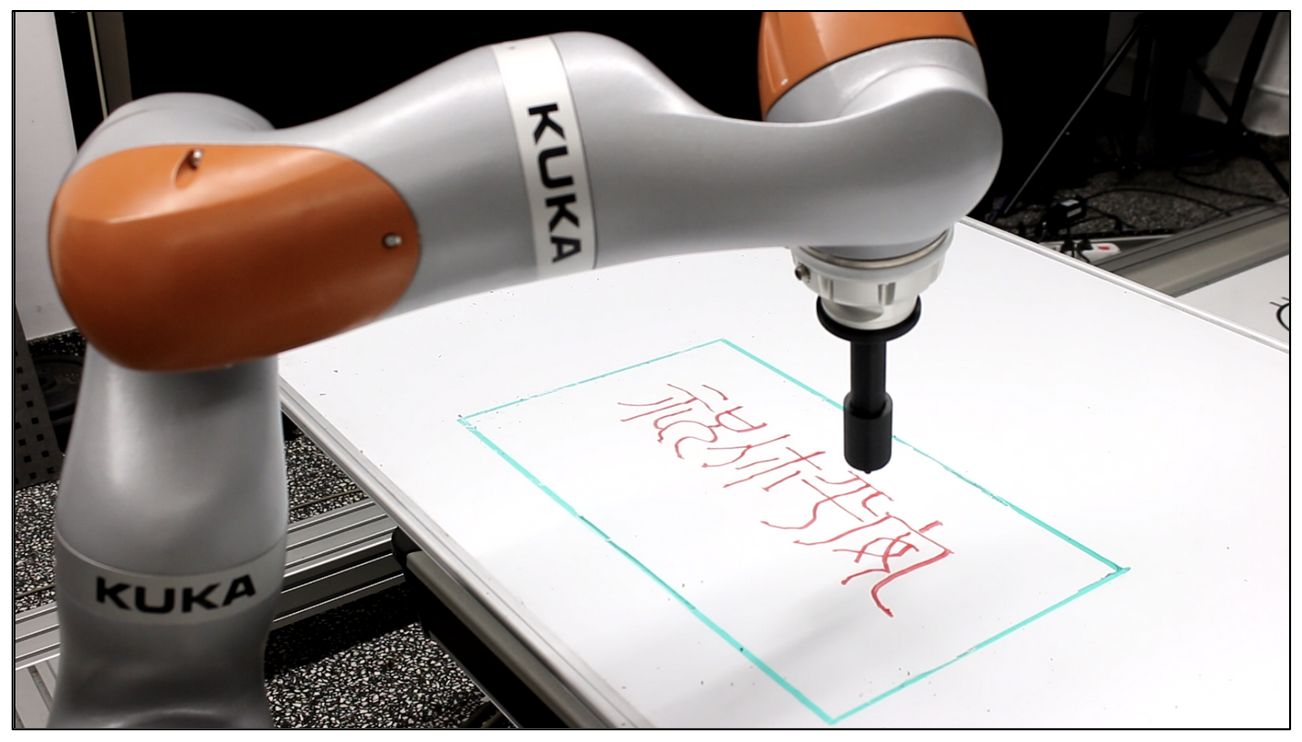}
    \caption{The setup of the robot drawing platform.}
    \label{fig:robot_draw}
    \vspace{-1.5em}
\end{figure}

\subsection{Adaptive Friction Estimator}\label{sec:adap_est}
Following \cite{slotine1987adaptive}, we design an adaptive friction estimator based on the CE principle, but instead of using the sliding
mode control design to account for model mismatch from the un-estimated parts of the dynamics, we exploit the structural properties of the EL mechanical systems \cite{munoz2022high} and propose a back-stepping controller.

Following \cite{slotine1987adaptive}, we define the Lyapunov function as
\begin{equation*}
    V_0(t)=\frac{1}{2}\mathbf{s}^T\mathbf{M}(\mathbf{q})\mathbf{s}+\frac{1}{2}\tilde{\boldsymbol{\pi}}_f^T\Gamma_f^{-1}\tilde{\boldsymbol{\pi}}_f,
\end{equation*}
with $\mathbf{s}=\dot{\tilde{\mathbf{q}}} + \Sigma \tilde{\mathbf{q}}$ the sliding surface , $\tilde{\mathbf{q}}$, $\dot{\tilde{\mathbf{q}}}$ the position and velocity tracking error,  $\Gamma_f$ the gain matrix for friction estimation, $\Sigma$ the gain matrix to define the sliding surface. Under the following control law
\begin{equation}\label{eq:nomianl_ctrl}
    \tau=\mathbf{Y}\hat{\boldsymbol{\pi}}+\mathbf{Y}_f \hat{\boldsymbol{\pi}}_f-\mathbf{K}_D\mathbf{s},
\end{equation}
and the parameter updating law 
\begin{equation}\label{eq:param_update_law}
    \dot{\hat{{\boldsymbol{\pi}}}}_f=-\Gamma_f\mathbf{Y}_f^T\mathbf{s},
\end{equation}
the derivative of the Lyapunov function
    $\dot{V}_0(t) 
    =-\mathbf{s}^T(\mathbf{K}_D \mathbf{s} - \mathbf{Y}\tilde{\boldsymbol{\pi}})$.
In \cite{slotine1987adaptive}, the authors propose an augmented sliding mode controller to guarantee $\dot{V}_0\leq 0$. In this paper we shows that with the boundedness property of the regressor of EL mechanical systems, namely \textit{proposition 1} in Sec. \ref{sec:structural_properties}, and assuming the model mismatch from dynamic parameters is bounded as $|\tilde{\boldsymbol{\pi}}| < C$, we have $|\mathbf{Y}\tilde{\boldsymbol{\pi}}| < D$,
with a positive constant $D$. It's easy to see that if we set $\mathbf{K}_D$ to be sufficiently high, $\dot{V}_0\leq 0$ and the stability of the system is guaranteed.

The above high gain adaptive controller has non-zero steady-state error due to external disturbance which could cause biased estimation of friction, hence we propose a back-stepping controller as follows: with an integrator $\mathbf{\epsilon}$ augmented into the controller in Eqn. \ref{eq:nomianl_ctrl}, the backstepping control law $
\mathbf{u}$ and the parameter updating law in Eqn. \ref{eq:param_update_law}
\begin{align*}
    &\tau=\mathbf{Y}\hat{\boldsymbol{\pi}}+\mathbf{Y}_f \hat{\boldsymbol{\pi}}_f-\mathbf{K}_D\mathbf{s} +\epsilon \\
    &\dot{\mathbf{\epsilon}}          =\mathbf{u}, \quad \quad
     \dot{\hat{{\boldsymbol{\pi}}}}_f =-\Gamma_f\mathbf{Y}_f^T\mathbf{s}.
\end{align*}
With the Lyapunov function $V_0$, we construct the following augmented Lyapunov function
\begin{equation*}
    V_1(t) = V_0(t) + \frac{1}{2}\epsilon^T\Gamma_e^{-1}\epsilon,
\end{equation*}
with the gain matrix $\Gamma_e$, the derivative of $V_1$ is
\begin{align*}
    \dot{V}_1(t) 
                 & = -\mathbf{s}^T(\mathbf{K}_D\mathbf{s} + \mathbf{Y}\tilde{\boldsymbol{\pi}}) + \mathbf{s}^T\epsilon + \epsilon^T\Gamma_e^{-1}\mathbf{u}
\end{align*}
Taking $\mathbf{u}=-\Gamma_e\mathbf{s}$, $\dot{V}_1=-\mathbf{s}^T(\mathbf{K}_D\mathbf{s} + \mathbf{Y}\tilde{\boldsymbol{\pi}})\leq 0$, the stability of the closed-loop system is guaranteed.

\subsection{Robust Guarantees of the Adaptation Process}\label{sec:robust_adaptation}
In this section we highlight the input-to-state (ISS) stability nature of the proposed adaptive friction estimator and the passivity property of the friction phenomenon \cite{olsson1996observer}. Based on ISS and passivity, we analyze some strategies for robusifying the adaptive estimation process.

Following the above analysis, the derivative of the  Lyapunov function of $V_1$ satisfies
\begin{align*}
    \dot{V}_1(t) & = -\mathbf{s}^T(\mathbf{K}_D\mathbf{s} - \mathbf{Y}\tilde{\boldsymbol{\pi}})   \quad                     
                \leq -\mathbf{s}^T\mathbf{K}_D\mathbf{s} + |\mathbf{s}||\mathbf{Y}||\tilde{\boldsymbol{\pi}}|                                                                                     \\
                 & \leq -\mathbf{s}^T\mathbf{K}_D\mathbf{s} + c |\tilde{\boldsymbol{\pi}}|    \ \quad                               \leq -\alpha(|\tilde{\mathbf{x}}|)+ \beta(|\tilde{\boldsymbol{\pi}}|),
\end{align*}
with $|\mathbf{Y}|$ the induced norm of the regressor matrix, $c=\sup(|\mathbf{s}|)|\mathbf{Y}|$ a positive constant, 
$\tilde{\mathbf{x}}=\begin{bmatrix}
        \tilde{\mathbf{q}} & \dot{\tilde{\mathbf{q}}}
    \end{bmatrix}^T$ the system's tracking error, and the class $\mathcal{K}_
    \infty$ functions \cite{agrachev2008input} $\alpha$  and $\beta$. The last inequality is easy to check by noticing that $c |\tilde{\boldsymbol{\pi}}|$ is a class $\mathcal{K}_\infty$ function and by scaling the $l_2$ norm of $|\tilde{\mathbf{x}}|$ with a positive constant $k$ we can have $k|\tilde{\mathbf{x}}|>|\mathbf{s}|$. Hence linear scaling with the rate $k$ can be a candidate $\alpha$ function. The last inequality establishes ISS, namely the tracking error is bounded by parameter variations. In the derivation, the second inequality comes from the boundedness of the regressor matrix and sliding mode tracking error due to stability.

From the second inequality, it is suggested to use configurations that have a small induced norm of the regressor matrix to reduce the disturbance.
\finalSentence{In this paper, we compensate for the friction between the joints, which depends on the relative motion of two adjacent joints. The friction models between different joints can then be considered decoupled; we can estimate friction parameters for each joint independently.}
Using the Fourier parameterized trajectories in \cite{sturz2017parameter}, we propose an optimization problem as follows for excitation generation
\begin{align}\label{eq:opt_traj}
    \min_{a_{j,i}, b_{j,i}}\ &\text{cond}(\mathbf{Y}(\mathbf{q}, \dot{\mathbf{q}}, \ddot{\mathbf{q}})) \\
    \text{s.t.}\ &q_j(t) = \sum_{i=1}^{N}\frac{a_{j,i}}{iw}\sin(i\omega t) - \frac{b_{j,i}}{iw}\cos(i\omega t) + q_j(0), \nonumber\\
                 &q_{j}^-(t) = q_{j, \text{fixed}}^-, \rebuttalThree{\ q_{\min} < q_j < q_{\max}, \ \dot{q}_{\min} <\dot{q}_j < \dot{q}_{\max}}, \nonumber 
\end{align}
with $j\in [1,J]$, the joint index of the robot, $i\in [1, N]$, the order for the fourier series, $q_j^-$ the joints that stay at fixed positions to decouple the system dynamics during estimation, 
\rebuttalThree{$q_{\min}$, $q_{\max}$, $\dot{q}_{\min}$, $\dot{q}_{\max}$, the joint position and the velocity limits, $\text{cond}$, the condition number, $\text{cond}(\mathbf{Y})=\frac{\sigma_{\max}(\mathbf{Y})}{\sigma_{\min}(\mathbf{Y})}$, where $\sigma_{\min}$ and $\sigma_{\max}$ are the minimum and maximum singular values of the regressor matrix.}
Instead of minimizing the induced norm of the regressor matrix, we minimize the condition number of the regressor matrix, which has a similar effect.
The optimization problem can be solved by sequential least square quadratic programming (SLSQP) or trust-region algorithms for constrained optimization
(trust-constr). 

Passivity is an inherent property of friction \cite{olsson1996observer}, to guarantee the passivity of the friction model, all parameters in the proposed friction model $\boldsymbol{\pi}_f$ must be positive. In this paper we simply clip the parameters $\hat{{\boldsymbol{\pi}}}_f = \text{clip}(\hat{{\boldsymbol{\pi}}}_f, 0, +\infty)$
after each update to guarantee the passivity of the friction model.

\section{EXPERIMENTAL EVALUATION}
In this section, we present experiment results for excitation generation, parameter estimation and evaluation. 
First, we conduct experiments to generate excitations for friction estimation. Then, we estimate friction with our proposed adaptive controller and the generated excitation. Subsequently, we conduct experiments to evaluate the estimated friction model with different nominal controllers, including PD, PID, and ADRC \cite{cheng2019active}, with random Fourier trajectories. For PD and PID controllers, we use an extended state observer \cite{cheng2019active} to estimate the velocity. 

\subsection{Experimental Setup}
We perform experiments using the Kuka LBR IIWA 14 manipulator. \rebuttalOne{The robot is a 7 DoF robot that has 7 revolute joints} and is controlled by a \rebuttalThree{real-time torque} control system with a sampling rate of $1kHz$. \rebuttalThree{The robot has a safety mechanism only triggers when the robot is beyond joint position, velocity or torque limits.} We designed a 3D-printed penholder as an end-effector for the drawing task. The drawing platform consists of the end-effector and a whiteboard in front of the robot, shown in Fig. \ref{fig:robot_draw}. 
 To reduce the effect of friction from contacts, we calibrate the pen tip and add a 2 cm spring so that the pen tip can have as slight contact with the whiteboard as possible. In this paper, when estimating the friction parameters and validating the estimated friction models with Fourier series trajectories, we use the robot without the end-effector; while performing the drawing tasks, we mount the end-effector on the robot.
 \rebuttalOne{The controller gains, $\mathbf{K}_D$ and $\Sigma$ are set based on the bandwidth-parameterization approach \cite{gao2003scaling} to simulate the controller as a critically damped system.
$\Gamma_f$ is set to achieve fast parameter convergence while maintaining a robust update. The backstepping gain $\Gamma_e$ is set to achieve a faster convergence rate to remove the non-zero steady-state error, we choose a gain of 10 for this paper.}

\begin{figure*}[tb!]
     \centering
     \begin{subfigure}[t]{0.3\textwidth}
         \centering
         \includegraphics[width=\textwidth]{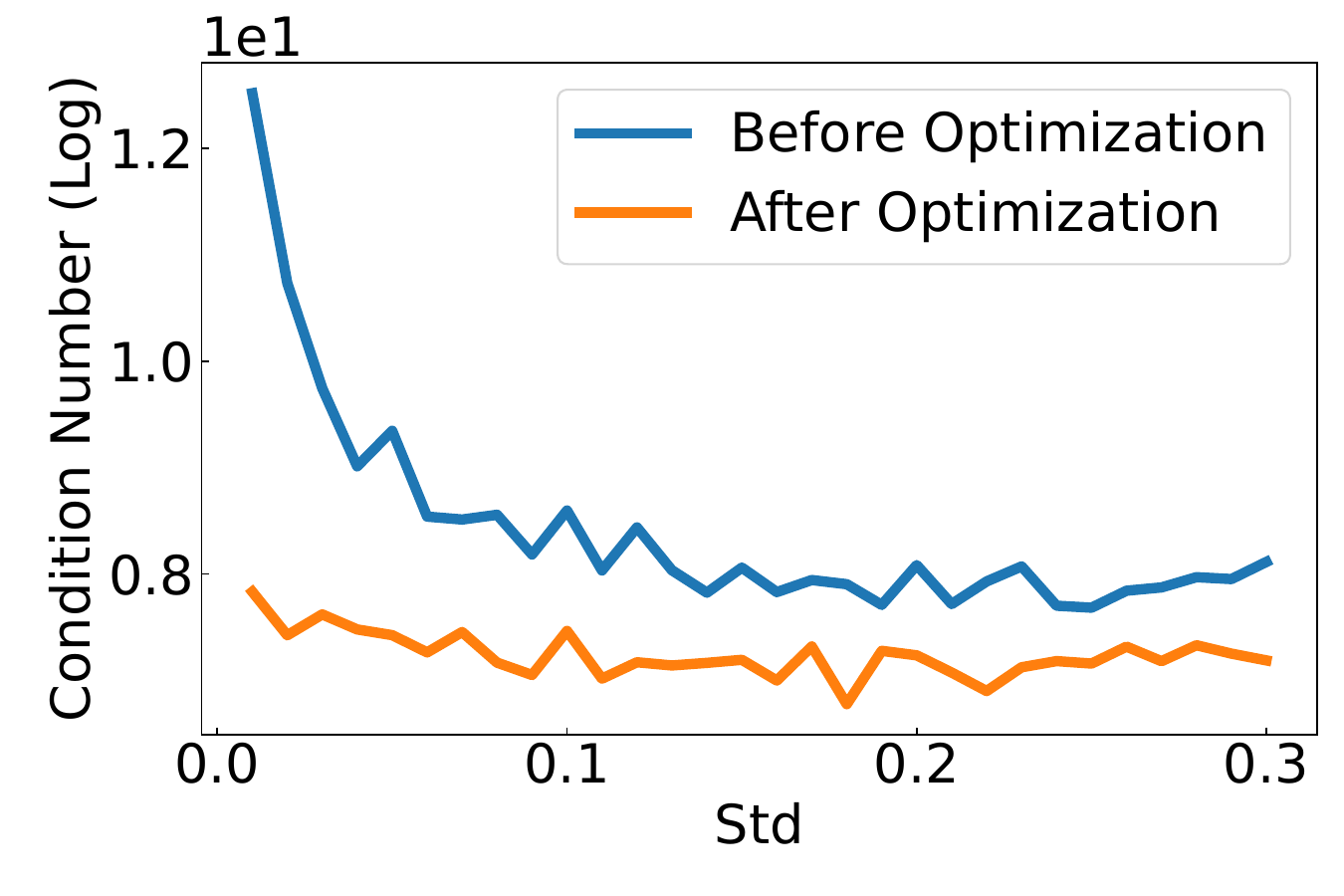}
         \caption{The effect of std of sampling distribution on initial and optimized excitations.}
         \label{fig:sampling_distribution_condtion_number}
     \end{subfigure}
      \hfill
     \begin{subfigure}[t]{0.3\textwidth}
         \centering
         \includegraphics[trim={0 0.1cm 0 0.1cm}, width=\textwidth]{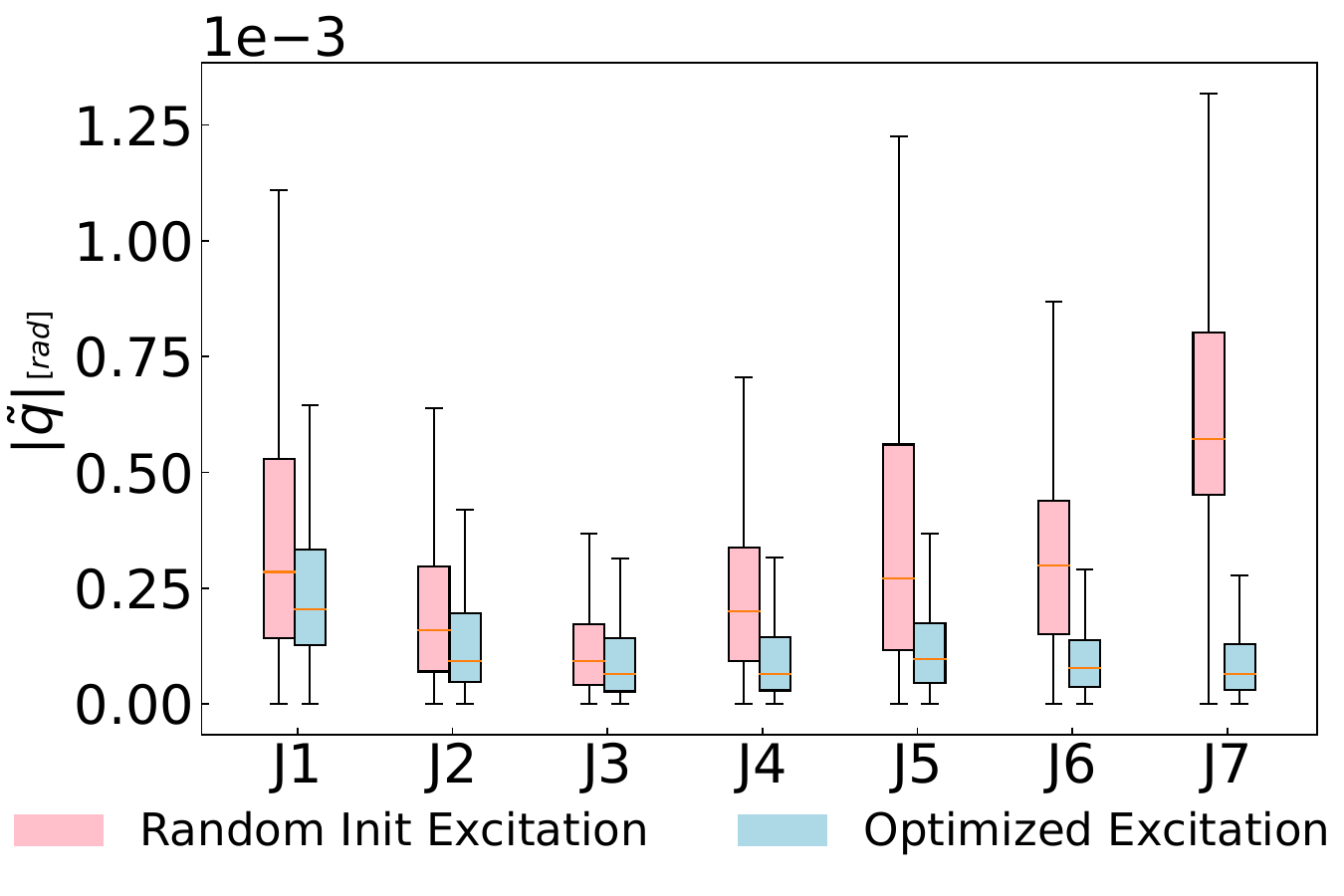}
         \caption{Position tracking error for estimation with the initial and optimized excitations.}
         \label{fig:excitation_comparison}
     \end{subfigure}
     \hfill
     \begin{subfigure}[t]{0.3\textwidth}
         \centering
         \includegraphics[trim={0 0.1cm 0 0.1cm}, width=\textwidth]{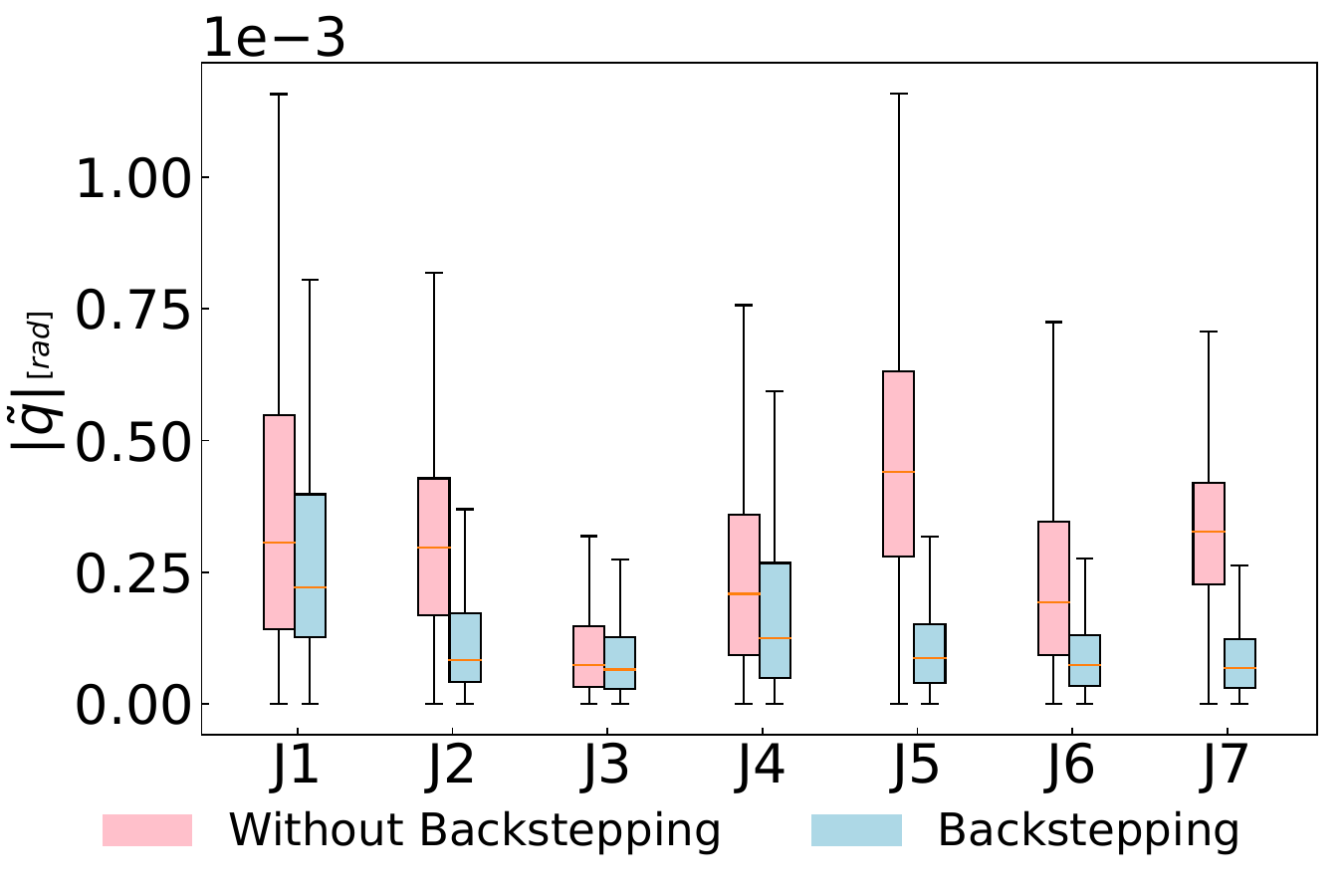}
         \caption{Comparison of the backstepping and non-backstepping design.}
         \label{fig:backstepping}
     \end{subfigure}
    \caption{The effect of excitation generation and backstepping design. (\textbf{Left}) the effect of std to the condition number of initial (blue) and optimized (orange) trajectories. (\textbf{Middle}) the position tracking error for estimation with initial and optimized excitations. (\textbf{Right}) the position tracking error for estimation compare with the backstepping and non-backstepping design.}
    \label{fig:three graphs}
    \vspace{-1.5em}
\end{figure*}

    


\begin{figure}[b]
    \vspace{-1.5em}
    \centering
    \includegraphics[width=\columnwidth]{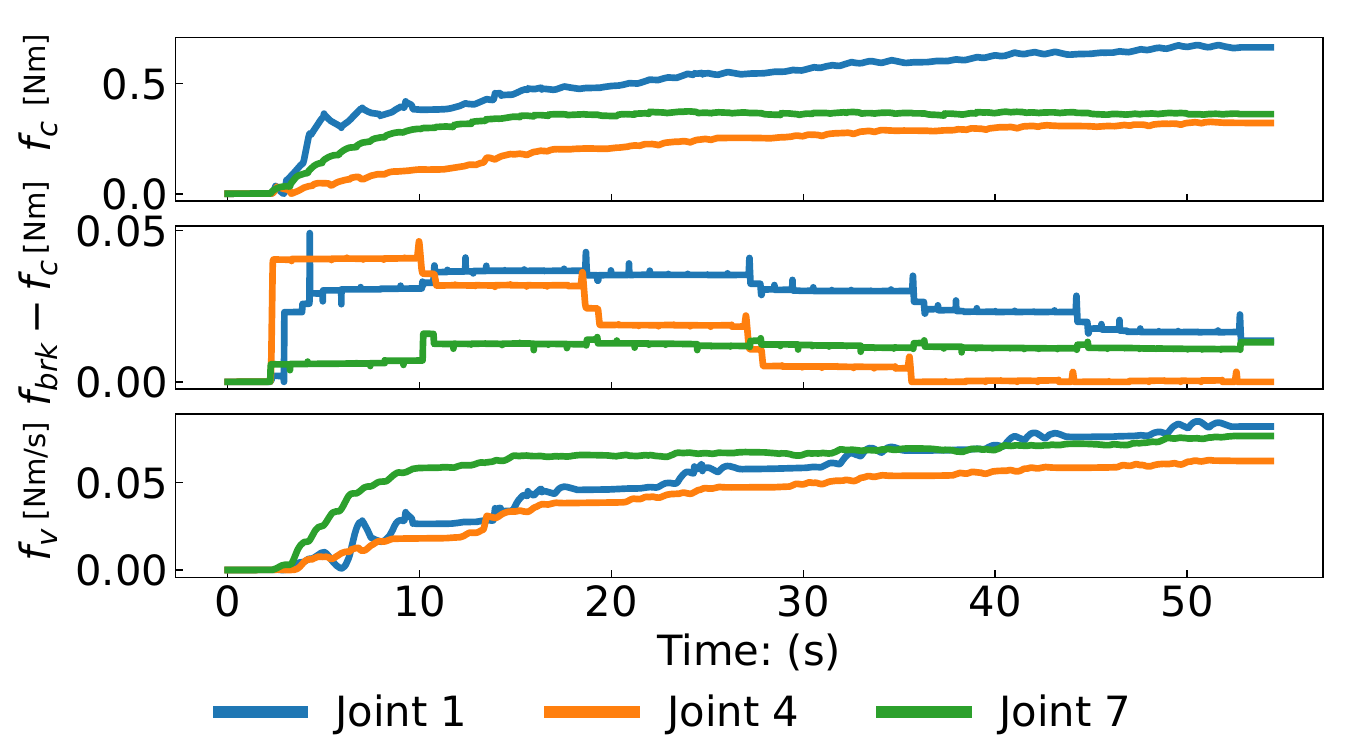}
    \caption{Parameter convergence in estimation. The parameters are from joint 1, 4, 7. From top to bottom, the coulomb friction $f_c$, viscous friction $f_v$, and the difference between breakaway force and coulomb force $f_{brk}-f_c$ are shown. }
    \label{fig:estimation_friction}
\end{figure}


\subsection{Excitation Generation and Estimation}
The excitation generation problem defined in Eqn. \ref{eq:opt_traj} is solved with the SLSQP solver available in scipy \cite{virtanen2020scipy}. \rebuttalThree{The generated excitations are within a safe region due to the constraints of joint position and velocity, which allows us to validate our approach without triggering the robot's safety mechanism.} The initial parameters of the trajectories are sampled from a Gaussian distribution with a mean of 0 and a standard deviation (std) of 0.05. \rebuttalSix{The effect of std on the excitation generation is studied, and results are shown in Fig. \ref{fig:sampling_distribution_condtion_number}. In Fig. \ref{fig:sampling_distribution_condtion_number}, the condition number of the initial trajectories decreases as std increases until 0.15. Moreover, since std does not affect the condition number of the optimized excitation, the selection of 0.05 is reasonable.}

Furthermore, the condition number of the regressor matrix, a key indicator of numerical stability and sensitivity to disturbances, is considerably lower for the optimized trajectory. Specifically, the condition number for the initial random trajectory is 5898, while for the optimized trajectory, it is reduced to 1819. This reduction by more than threefold indicates that the optimized trajectories can substantially mitigate the impact of model mismatch and lead to more reliable estimation results.
\finalSentence{Moreover, the excitation is generated offline which makes the algorithm practical. }

The convergence of friction parameters during the estimation process is shown in Fig. \ref{fig:estimation_friction}. For clarity, we present a selection of parameters from joints 1, 4, and 7. These parameters include Coulomb friction $f_c$, viscous friction $f_v$, and the difference between the breakaway force and Coulomb force $f_{brk} - f_c$. As illustrated in the figure, all three parameters converge to a small region, which demonstrates the convergence of the proposed adaptive controller. \rebuttalSix{The convergence time takes 50 seconds due to the usage of slow varying trajectories and the parameter updating gain we set considers not only faster parameter convergence but also robustness to avoid chattering updates in parameters.}

The convergence of friction parameters is critical as it indicates that the controller can adaptively tune the system to account for varying friction characteristics, ensuring consistent and reliable performance. The small region of convergence suggests that the adaptive controller is stable across different operating conditions, \rebuttalOne{including joint velocities and payloads.} This behavior is vital in systems where even minor discrepancies in friction estimation could lead to significant errors in performance.

\begin{figure*}[!tb]
    \centering
    \includegraphics[trim={0 0.4cm 0 0cm},width=\textwidth]{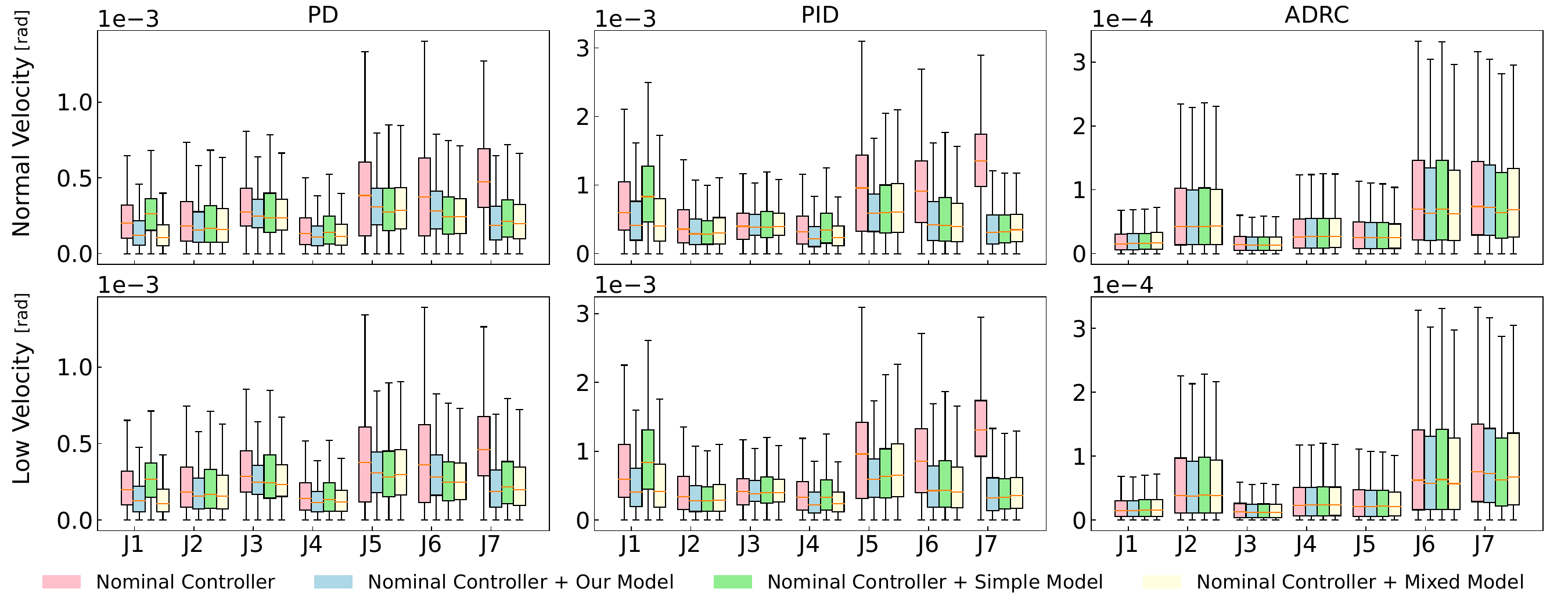}
    \caption{
        Evaluation results with Fourier trajectories. The first row compares the absolute joint position tracking error for nominal controllers. The second row shows the same comparison, but under low-speed conditions $|\dot{q}_d|<0.01$. In each sub-figure, from left to right shows the result from joint 1 to 7. In each box plot, the orange line represents the median value, the box represents the interquartile range, and the whiskers represent the range of the data.
    }
    \label{fig:fourier_trajectory_ctrls}
    \vspace{-1.5em}
\end{figure*}

\begin{figure}[t]
    \centering
    \includegraphics[width=0.9\columnwidth]{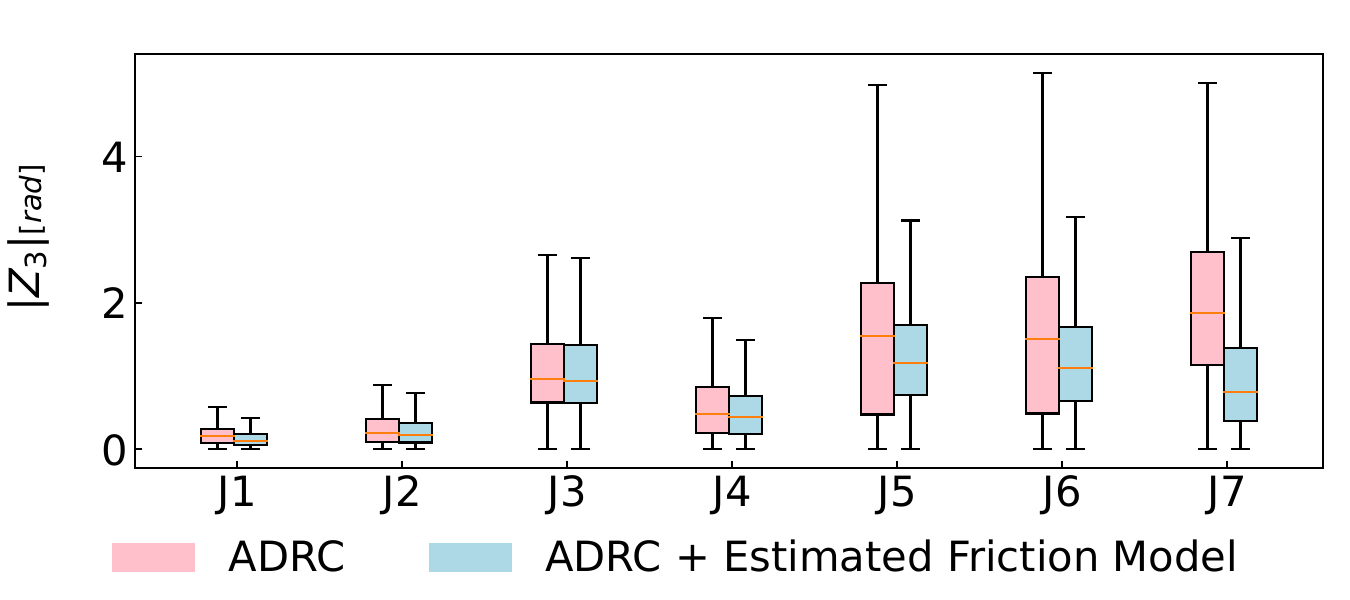}
    \caption{The disturbance estimation from ADRC with and without the estimated friction model.}
    \label{fig:eval_z3}
    \vspace{-2.0em}
\end{figure}

To demonstrate the necessity of using optimized excitation, we conducted experiments comparing the tracking errors between the initial random trajectories and the optimized trajectories during the friction estimation process. The results, presented in Fig. \ref{fig:excitation_comparison}, clearly show that the position tracking error for the optimized trajectories is significantly smaller than that of the initial random trajectories across all joints. The reduction in tracking error highlights the effectiveness of the optimized trajectories in minimizing disturbances and improving the accuracy of friction estimation.

\rebuttalOne{The effect of backstepping is also investigated, and results are shown in Fig. \ref{fig:backstepping}. The results show that backstepping leads to a more zero-centered tracking error compared with the non-backstepping one, indicating that the backstepping design can reduce biased estimation and potential chattering.}

\subsection{Validation with Fourier Trajectories}
During the validation phase, our goal is to evaluate the robot's static and dynamic performance with the estimated friction model. Static performance is characterized by tracking errors that occur under low-velocity conditions, in this paper we defined as the absolute value of the desired joint velocity falls below $0.01$. We compare the absolute joint position tracking errors with different nominal controllers to provide a comprehensive evaluation.

The trajectories we use for validation are random Fourier-series trajectories from \cite{sturz2017parameter}. We collect a total of 10 unique trajectories of 100 seconds, with each trajectory being repeated ten times. To avoid biased evaluation due to the different initial states of the robot, we reset the robot three times to evaluate each trajectory, so overall, we evaluated the model with a 30000-second dataset. Since Fourier series trajectories can cover a broad spectrum of frequencies, they allow us to evaluate the estimated friction model's static and dynamic performance. We mainly focus on slow-varying trajectories with many zero-crossing velocities, so the friction phenomenon happens more frequently.

The results for the Fourier trajectories are depicted in Fig. \ref{fig:fourier_trajectory_ctrls}. \rebuttalSix{The baselines for evaluation include: a) no friction model; b) our estimated model; c) a simplified model that neglect the nonlinear effect, defined as $\tau_f=f_c\tanh(\frac{v}{v_{\text{coul}}})+f_{\text{vis}}v$; d) a mixed model setting: using our model for joints 1,2,3, and 4, and the simplified model for joints 5, 6, and 7.} A detailed analysis of the mean value, interquartile range, and data range of position tracking errors reveals that incorporating the estimated friction model leads to a significant reduction in absolute position tracking errors when compared to the nominal PD and PID controllers. The improvement is consistently observed across both types of controllers, demonstrating the broad effectiveness of incorporating the estimated friction model into different control strategies. \rebuttalSix{The comparison of our model and the simplified model shows a significant performance gain, especially for joints 1, 2, 3, and 4. 
The mixed model setting, which reaches the best controller performance, shows that our model can be used to improve the overall controller's performance even further.}

Although the ADRC (Active Disturbance Rejection Control) group shows no significant improvement in absolute position tracking errors with the estimated friction model compared to the nominal ADRC setup, other benefits become apparent. \rebuttalSix{We investigated the disturbance reduction perspective of our proposed model with the extended state observer.} As illustrated in Fig. \ref{fig:eval_z3}, the integration of the friction model enhances the performance of the extended state observer by reducing disturbance estimation. The improvement leads to a more accurate representation of system disturbances, enabling a safer and more compliant ADRC controller design. The friction model's ability to refine disturbance estimation contributes to better overall system performance and robustness, highlighting its valuable role in enhancing control strategies beyond just reducing tracking errors.

Furthermore, the tracking error under low-velocity conditions clearly shows the advantages of nominal controllers equipped with the friction model. This indicates that the model effectively compensates for dynamic and static friction phenomena. By improving performance even at slow speeds, where static friction effects are more pronounced, the model demonstrates its ability to enhance control accuracy across a range of operational conditions. \rebuttalSix{We further compare our method with the state-of-the-art controllers \cite{dietrich2021practical, cheng2019active}. The results in Table \ref{table:comparison_state_of_the_art} show our controller outperforms the state-of-the-art controllers significantly.}

\begin{table}[b]
\centering
    \vspace{-2em}
    \begin{tabular}{ |c| c |c| c| } 
      \hline
      \textbf{Method} & \textbf{DoF} & \textbf{Error Scale} \\ 
      \hline
      ADRC + Our Model & 7 & e-4 \\ 
      \hline
      Inverse Dynamics \cite{dietrich2021practical} & 7 & e-2 \\ 
      \hline
      ADRC + Feedforward \cite{cheng2019active} & 6 & e-3 \\
      \hline
    \end{tabular}
    \caption{Comparison of state-of-the-art controllers.}
    \label{table:comparison_state_of_the_art}
\end{table}

\begin{figure}[t]
    \centering
    \includegraphics[width=0.85\columnwidth]{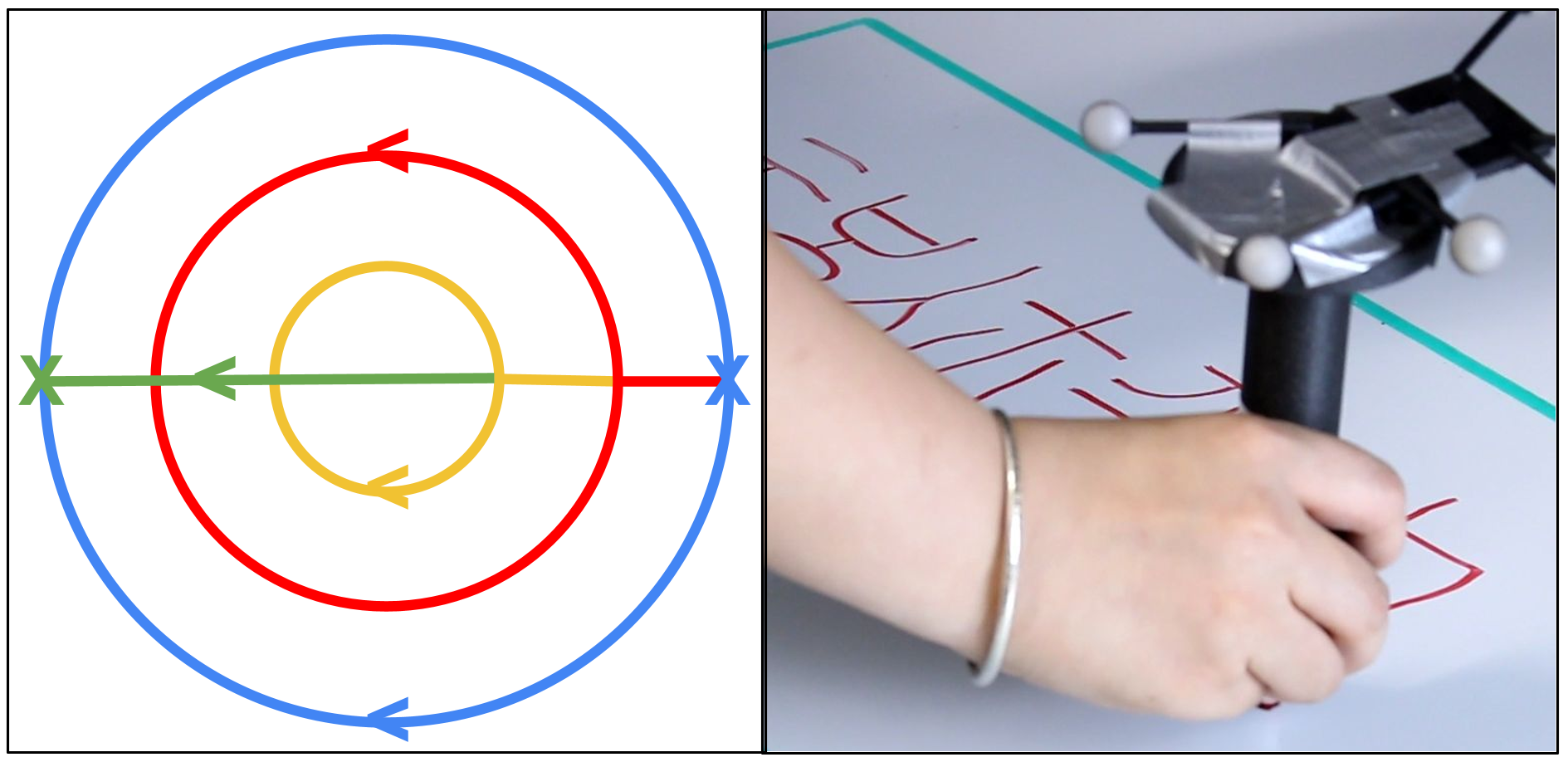}
    \caption{Trajectory generation for the drawing tasks. (\textbf{Left}) the circular motion; (\textbf{Right}) the data collection of Chinese characters. The circular motion starts at the blue cross and ends at the green one. It's composed of four phases, denoted by four different colors.}
    \label{fig:eval_draw_data_generation}
    \vspace{-1.5em}
\end{figure}

\begin{figure}[b]
    \vspace{-2em}
    \centering
    \includegraphics[width=1.0\columnwidth]{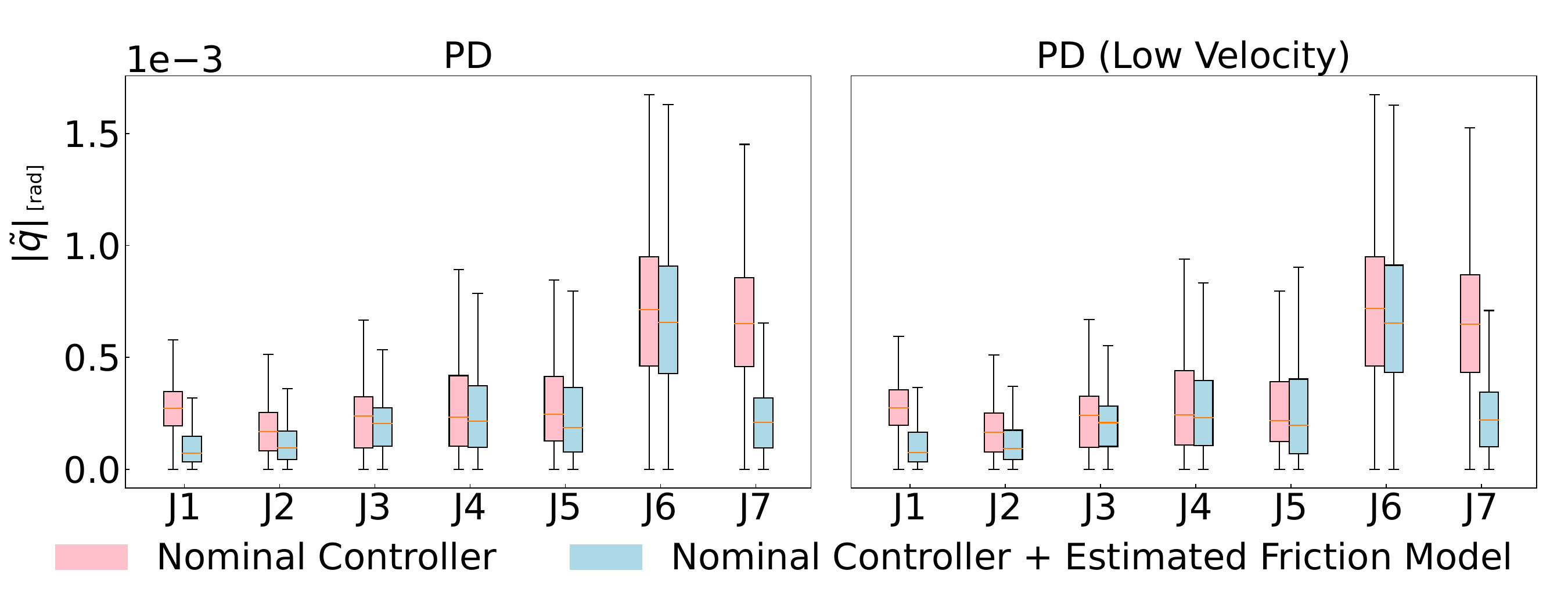}
    \caption{Evaluation result of the circular motion. (\textbf{Left}) results under normal trajectory (\textbf{Right}) results with the filtered low velocity. In each sub-figure, from left to right, shows the comparison between PD and PD with the estimated model.}
    \label{fig:eval_circle_tracking_error}
\end{figure}

\subsection{Path Drawing Validation}
In this section, we present the results of the trajectory tracking experiments for drawing tasks in Cartesian space with two types of trajectories. The first one is the circular trajectory, which combines a closed curve, a circle, and straight lines. This trajectory offers a well-rounded test of the robot's tracking capabilities. It allows us to assess how effectively the system can handle smooth and continuous motion paths.
On top of the circular motion trajectory, we collect Chinese character trajectories for more complicated trajectory tracking. The Chinese character trajectory involves intricate patterns with more frequent turns and sharp corners. 
The evaluation of the two drawing tasks are under the same setup: a nominal PD controller without the estimated friction model and a PD controller with the estimated friction model.

\paragraph{Evaluation of Circular path drawing}
The left figure in Fig. \ref{fig:eval_draw_data_generation} illustrates the circular motion trajectory. From Fig. \ref{fig:eval_draw_data_generation}, the circular motion trajectory consists of four phases, and the varying colors indicate the change of phases. The rotation of the circular motion is denoted by the arrows.

\begin{figure}[t]
    \centering
    \includegraphics[trim={0 -0cm 0 0cm}, width=0.9\columnwidth]{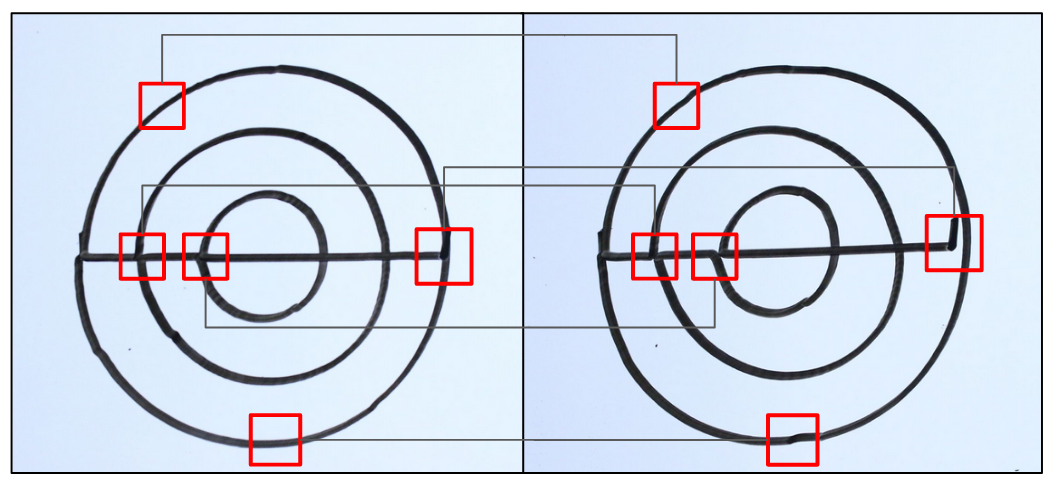}
    \caption{Evaluation result of the circular motion. (\textbf{Left}) the result of a PD controller with the estimated friction model; (\textbf{Right}) the result of a PD controller. Red boxes denote the key difference of the two figures.}
    \label{fig:eval_circle_drawing}
    \vspace{-2.0em}
\end{figure}

The evaluation of the circular motion includes quantitative and qualitative studies. First, we repeat the drawing tasks 20 times and compare the absolute position tracking error for the PD controller and PD with the estimated friction model, shown in Fig. \ref{fig:eval_circle_tracking_error}. Then, we move on to the evaluation in cartesian space to validate the estimated model's ability to improve the circular path drawing, shown in Fig. \ref{fig:eval_circle_drawing}. In Fig. \ref{fig:eval_circle_tracking_error}, it is clear that the group with the estimated friction model significantly reduces the tracking error in normal and low-velocity settings, which highlights the versatility of the estimated model for different trajectories. As shown in Fig. \ref{fig:eval_circle_drawing}, the drawing of the closed curve is more accurate, and the shape of the circle is notably rounder. The red boxes in the figure emphasize the enhanced detail and precision achieved with the friction model, demonstrating its impact on improving trajectory adherence and reducing distortions.

\paragraph{Qualitative results of Calligraphy Writing}
\begin{figure}[b]
    \centering
    \includegraphics[trim={0 0cm 0 0.8cm}, width=0.8\columnwidth]{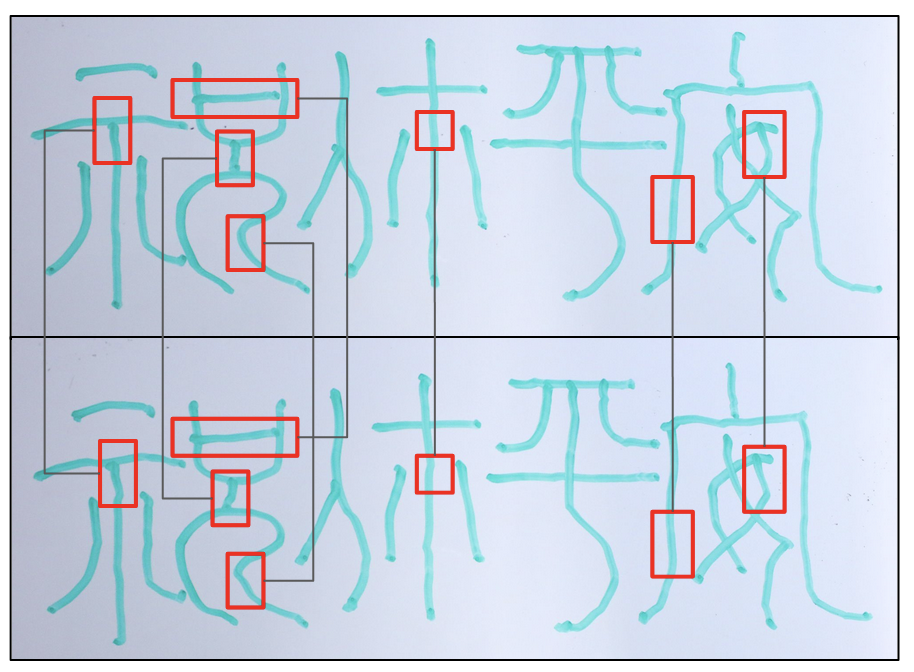}
    \caption{Evaluation result of drawing Chinese characters. (\textbf{Top}) the result of a PD controller with the estimated friction model; (\textbf{Bottom}) the result of a PD controller. Red boxes denote the key difference of the two figures.}
    \label{fig:eval_characters_drawing}
\end{figure}
The data collection process of the calligraphy writing task is depicted on the right side of Fig. \ref{fig:eval_draw_data_generation}. Here, a human demonstrator gathers data using the OptiTrack system, which captures precise motion information. 
The raw Cartesian space trajectories obtained from the OptiTrack system are then converted into joint space trajectories using inverse kinematics.
A Savitzky-Golay filter is applied to smooth the trajectories and obtain velocities and accelerations. We chose four Chinese characters that translate to "hope you are safe" in English, rendered in the \textit{small seal script} drawing style. The choice of this drawing style introduces greater complexity due to the numerous turns and detailed shapes of each character.

The qualitative results of the calligraphy writing were conducted in a more blinded manner, as illustrated in Fig. \ref{fig:eval_characters_drawing}. We first perform the drawing tasks with PD and PD equipped with the estimated friction model, then ask the human demonstrator to compare the two figures without knowledge of the controller settings. As highlighted by the red boxes in the figure, the demonstrator observed that the strokes of the Chinese characters were smoother, and the corners were sharper with the estimated friction model. This feedback underscores the practical benefits of the friction model in handling more complex and detailed drawing tasks, further validating its effectiveness in enhancing the overall performance and accuracy of the robot's motion control.

\section{CONCLUSIONS}
In this paper, we propose a linear friction model combined with an adaptive friction estimator, leveraging the certainty equivalence principle and back-stepping techniques to estimate friction parameters effectively. We also introduce an excitation generation algorithm based on input-to-state stability. The linear friction model facilitates the design of certainty equivalence-based adaptive controllers, which are shown to be effective in practice. The adaptive friction estimator, paired with the excitation algorithm, enables robust and accurate estimation of friction parameters, as demonstrated by our experimental results. These results highlight significant improvements in static and dynamic performance for the robot manipulator, showcasing the effectiveness and robustness of our proposed methods. 
\rebuttalOne{Our proposed method has the potential to be used in tracking tasks that require proper friction compensation such as grinding or polishing.}

However, while adaptive controllers offer advantages by eliminating the need for joint torque measurements, the certainty equivalence design does not guarantee parameter convergence. Another limitation of this work is the uncertainty of the friction model is not modeled, which prevents the usage of the model in completely unseen scenarios. 
\rebuttalOne{Moreover, our model does not consider time-variant factors, e.g., temperatures, so it can not capture changes in non-stationary environments. However, our adaptive design allows online adaptation to time variant changes, additional ablation study can be found on the project website.}
Future work will focus on characterizing the uncertainty associated with friction model parameters and
\rebuttalOne{include time-variant factors in friction modeling.}
Such advancements could enhance the reliability of the adaptive control strategy and extend its applicability to more complex and variable operating conditions.






\section*{ACKNOWLEDGMENT}
The support provided by China Scholarship Council (No. 202008440452) is acknowledged. 
Thanks Jiahui Shi and Weiyu Cheng for the help in drawing Chinese characters.

\bibliographystyle{plain} 
\bibliography{ref} 

\end{document}